\documentclass[runningheads,anonymous]{llncs}
\usepackage[T1]{fontenc}
\usepackage{graphicx}
\usepackage{hyperref}
\usepackage{booktabs}
\usepackage{amsmath}
\usepackage{multirow}
\usepackage{standalone}
\usepackage{tikz}
\usepackage{xcolor}

\usetikzlibrary{shapes.geometric, shapes.misc, arrows.meta, positioning, calc, shadows}

\begin{document}
\title{On the Stability of Prompt Ranking in Large Language Model Evaluation}
%
%

\author{Shaoshuai Du\inst{1}\orcidID{0009-0006-4991-3494} \and
Penghao Liang\inst{2}\orcidID{0009-0001-0501-1003} \and
Yixian Shen\inst{1}\orcidID{0000-0001-8447-872X} \and
Chuanqi Shi\inst{3}\orcidID{0000-0001-5249-7803} \and
Hang Zhang\inst{3}\orcidID{0009-0007-8455-7373} \and
Lun Wang\inst{4}\orcidID{0000-0002-0447-8701}} 

%
%

\institute{University of Amsterdam, Amsterdam, Netherlands \email{s.du@uva.nl} \and
Northeastern University, Boston, MA, USA  \and
University of California San Diego, La Jolla, CA, USA\textbf{}\and
Duke University, Durham, NC, USA
}
\maketitle              
\begin{abstract}
Prompt-based interaction has become a dominant paradigm for using large language models (LLMs), where multiple candidate prompts are evaluated and the top-ranked one is selected for downstream use. This workflow implicitly assumes that prompt rankings are stable under minor variations in evaluation conditions. In this paper, we systematically study prompt ranking stability under common sources of variability, including random seeds and limited evaluation subsets. Across three open-weight LLMs and two benchmark tasks, we find that while overall rank correlations are often moderate to high, the identity of the top-performing prompt frequently changes, leading to unreliable selection decisions. To address this issue, we propose a simple stability-aware selection strategy based on a lower confidence bound, which accounts for both performance and variance. Our results show that this approach improves robustness in unstable settings while remaining competitive in more stable regimes. These findings highlight the importance of accounting for evaluation uncertainty in prompt selection and LLM benchmarking.

\keywords{Prompt engineering \and Large language models \and Evaluation stability \and Ranking robustness}
\end{abstract}

\section{Introduction}
As AI models continue to advance, large language models (LLMs) are increasingly accessed through prompt-based interfaces, where task behavior is specified via natural language instructions rather than task-specific training~\cite{DBLP:journals/corr/abs-2502-15770/lunwang,DBLP:journals/corr/abs-2412-20061/yanxin,DBLP:conf/nips/BrownMRSKDNSSAA20/language}.
As a result, prompt design and prompt selection have become central to both research and deployment of LLM-based systems.
In many practical workflows, multiple candidate prompts are evaluated on a benchmark, ranked by performance, and a single ``best'' prompt is selected for downstream use.

A common but often implicit assumption underlying these practices is that prompt performance rankings are stable.
That is, a prompt that outperforms others under a given evaluation protocol is expected to remain superior under minor variations of evaluation conditions.
This assumption motivates widespread practices such as selecting prompts based on average accuracy or reporting a single top-performing prompt.

In practice, however, prompt evaluation is subject to multiple sources of variability.
Evaluation protocols often rely on limited evaluation budgets, subsampling of benchmark datasets, or different random seeds.
While prior work has studied output variability under stochastic decoding or sensitivity to few-shot examples, prompt evaluation itself is typically treated as deterministic.
In particular, the stability of prompt \emph{rankings}, rather than absolute scores, has received little systematic attention.

Importantly, instability in prompt rankings has direct practical implications.
When multiple prompts achieve similar average performance, even small fluctuations in evaluation scores can lead to changes in their relative ordering.
Such ranking instability may cause prompt selection decisions to be overly sensitive to evaluation noise, resulting in brittle or non-reproducible outcomes.

In this work, we challenge the assumption of stable prompt rankings.
Rather than focusing on absolute performance variance, we study how the relative ordering of prompts changes under realistic evaluation variability.
Specifically, we ask the following questions:
(1) How stable are prompt performance rankings across random seeds and evaluation subset sizes?
(2) How does ranking instability affect common prompt selection strategies?
(3) Can prompt selection be made more robust using simple stability-aware criteria?

To answer these questions, we conduct a systematic empirical study in which a fixed set of prompts is repeatedly evaluated under controlled variations of evaluation conditions.
Our results reveal substantial ranking instability, particularly for small evaluation subsets, and show that selecting prompts based solely on mean performance can lead to unreliable decisions.
At the same time, we demonstrate that incorporating stability considerations via a simple lower confidence bound criterion can improve robustness under noisy evaluation conditions while remaining competitive in more stable settings.

Our contributions are as follows:

(1) We provide the first systematic study of prompt ranking stability under realistic evaluation variability, including random seeds and limited evaluation budgets.

(2) We show that high rank correlation does not necessarily imply stable prompt selection, revealing a gap between global ranking agreement and decision-level consistency.

(3) We propose a simple stability-aware selection method based on lower confidence bounds, which improves robustness under noisy evaluation conditions without sacrificing performance in stable regimes.

\section{Related Work}

\subsection{Prompt Engineering and Prompt Sensitivity}

Prompt engineering has emerged as a key technique for controlling the behavior of large language models without additional training~\cite{DBLP:conf/nips/Wei0SBIXCLZ22/wei22chain,DBLP:conf/nips/KojimaGRMI22/kojima2022large}.
Prior work has explored instruction design, reasoning cues, output constraints, and prompt ensembles, demonstrating that prompt formulation can significantly affect model performance.
Several studies have also reported sensitivity of LLM outputs to prompt phrasing, formatting, and example selection.

\subsection{Evaluation Variability and Benchmark Robustness}
Evaluation variability has been studied in various contexts, including randomness in model initialization, stochastic decoding, and dataset subsampling~\cite{DBLP:conf/emnlp/DodgeGCSS19/dodge2019show,DBLP:conf/acl/ReichartDBS18/dror2018hitchhiker}.
Prior work has shown that benchmark scores can be sensitive to evaluation noise, particularly under limited data or stochastic settings.
Bootstrap methods and statistical significance testing have been proposed to quantify uncertainty in model evaluation.

In contrast to these studies, our focus is not on estimating confidence intervals for absolute scores, but on understanding how evaluation variability affects the \emph{relative ordering} of prompts.

\subsection{Ranking Stability and Model Selection}
While ranking stability has been studied in areas such as information retrieval and model selection, these works typically focus on model-level comparisons~\cite{DBLP:conf/sigir/Voorhees98variations}.
In contrast, we investigate ranking stability at the prompt level, where performance differences are often subtle and evaluation noise plays a larger role.
Several works have noted that rank-based decisions may be sensitive to noise, even when aggregate performance metrics appear stable~\cite{DBLP:conf/emnlp/ReimersG17reporting}.

Our work brings this perspective to prompt evaluation for LLMs.
To our knowledge, this is the first work that systematically studies prompt ranking stability under controlled evaluation variability and connects ranking instability to prompt selection robustness.

\section{Method}

In this section, we formalize prompt evaluation as a stochastic ranking problem, where evaluation variability induces randomness in prompt performance and ranking outcomes.
Under this formulation, prompt ranking is no longer deterministic but depends on the underlying sampling distribution, and prompt selection corresponds to identifying robust optima under uncertainty.

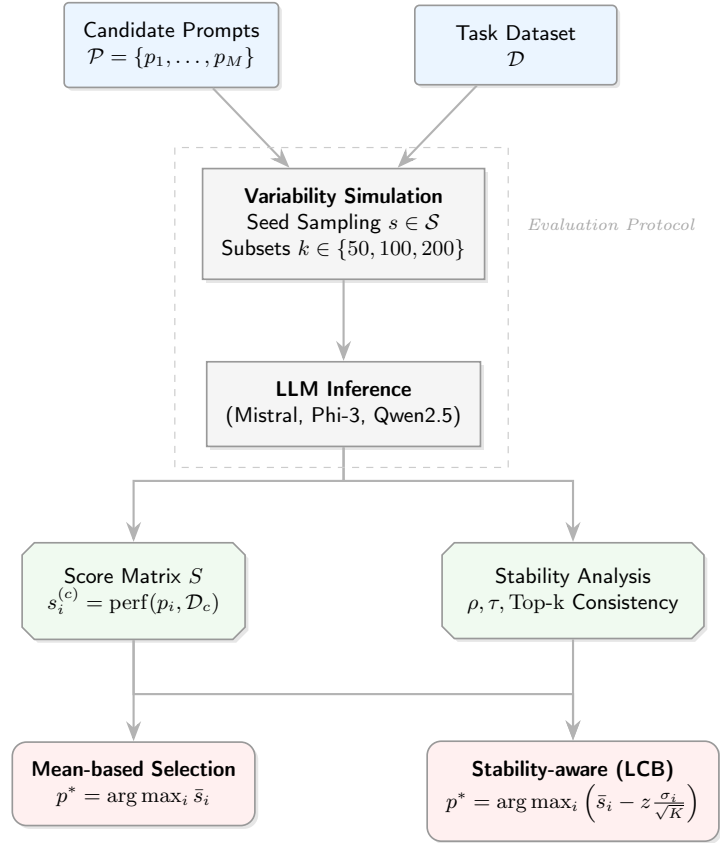
\begin{figure}[!h] 
\centering
\definecolor{inputblue}{RGB}{235,245,255}
\definecolor{processgray}{RGB}{245,245,245}
\definecolor{metricgreen}{RGB}{240,250,240}
\definecolor{resultred}{RGB}{255,240,240}

\begin{tikzpicture}[
    node distance=1.8cm,
    font=\sffamily\small,
    base/.style={
        draw=gray!80,
        line width=0.7pt,
        align=center,
        minimum height=1.1cm,
        inner sep=8pt,
        general shadow={
            fill=black,
            opacity=0.1,
            shadow xshift=0.5mm,
            shadow yshift=-0.5mm
        }
    },
    input/.style={base, fill=inputblue, rounded corners=2pt, minimum width=3cm},
    proc/.style={base, fill=processgray, minimum width=3.5cm},
    metric/.style={base, fill=metricgreen, chamfered rectangle, minimum width=3cm},
    final/.style={base, fill=resultred, rounded corners=6pt, font=\sffamily\bfseries, minimum width=3.5cm},
    arrow/.style={-{Stealth[scale=1.2]}, line width=0.9pt, draw=gray!60}
]

\node (prompts) [input] {Candidate Prompts \\ $\mathcal{P} = \{p_1, \dots, p_M\}$};
\node (dataset) [input, right=2cm of prompts] {Task Dataset \\ $\mathcal{D}$};

\node (sampling) [proc, below=1.2cm of $(prompts.south)!0.5!(dataset.south)$] {
    \textbf{Variability Simulation} \\
    Seed Sampling $s \in \mathcal{S}$ \\
    Subsets $k \in \{50, 100, 200\}$
};

\node (execution) [proc, below=1.2cm of sampling] {
    \textbf{LLM Inference} \\
    (Mistral, Phi-3, Qwen2.5)
};

\node (matrix) [metric, below left=1.5cm and -0.5cm of execution] {
    Score Matrix $S$ \\
    $s_i^{(c)} = \mathrm{perf}(p_i, \mathcal{D}_c)$
};
\node (stability) [metric, below right=1.5cm and -0.5cm of execution] {
    Stability Analysis \\
    $\rho, \tau, \mathrm{Top\text{-}k}$ Consistency
};

\node (mean_sel) [final, below=1.5cm of matrix] {
    Mean-based Selection \\
    $p^* = \arg\max_i \bar{s}_i$
};
\node (lcb_sel) [final, below=1.5cm of stability] {
    Stability-aware (LCB) \\
    $p^* = \arg\max_i \left(\bar{s}_i - z \frac{\sigma_i}{\sqrt{K}}\right)$
};

\draw [arrow] (prompts) -- (sampling);
\draw [arrow] (dataset) -- (sampling);
\draw [arrow] (sampling) -- (execution);
\draw [arrow] (execution.south) -- ++(0,-0.5) -| (matrix.north);
\draw [arrow] (execution.south) -- ++(0,-0.5) -| (stability.north);
\draw [arrow] (matrix) -- (mean_sel);
\draw [arrow] (stability) -- (lcb_sel);
\draw [arrow] (matrix.south) -- ++(0,-0.8) -| (lcb_sel.north);

\draw [dashed, gray!40, line width=0.5pt]
    ($(sampling.north west)+(-0.4,0.3)$) rectangle ($(execution.south east)+(0.4,-0.3)$);
\node [right=0.5cm of sampling, text=gray!70, font=\itshape\scriptsize] {Evaluation  Protocol};

\end{tikzpicture}
\caption{Overview of the proposed prompt evaluation and selection framework. The pipeline simulates evaluation variability through multi-seed subsampling before applying the stability-aware selection strategy.}
\label{fig:method_pipeline}
\end{figure}

\subsection{Problem Formulation}

Let $\mathcal{P} = \{p_1, p_2, \dots, p_M\}$ denote a fixed set of candidate prompts for a given task, and let $\mathcal{D}$ denote the evaluation dataset.
Under an evaluation condition $c$ (e.g., a specific random seed and subset of examples), each prompt $p_i$ is assigned a performance score $s_i^{(c)}$, such as accuracy.

This induces a ranking over prompts:
\[
\pi^{(c)} = \text{rank}\big(\{s_i^{(c)}\}_{i=1}^M\big),
\]
where lower ranks in the induced permutation correspond to better-performing prompts.

We consider multiple evaluation conditions $\mathcal{C} = \{c_1, \dots, c_K\}$, obtained by varying random seeds and evaluation subsets.
Our goal is to analyze how stable the rankings $\{\pi^{(c)}\}$ are across different conditions.

\subsection{Evaluation Variability}

We model evaluation variability as arising from two sources:

\textbf{Random seed variation:} Different seeds induce different subsamples of the dataset.

\textbf{Subset size variation:} We evaluate on subsets of size $k \in \{50, 100, 200\}$ to simulate limited evaluation budgets.

For each condition $c$, all prompts are evaluated on the same subset to ensure fair comparison.
This produces a score matrix of size $|\mathcal{C}| \times M$, where each row corresponds to a ranking over prompts.

\subsection{Ranking Stability Metrics}

To quantify the similarity between rankings, we compute pairwise correlations across evaluation conditions.

\paragraph{Rank Correlation.}
Given two rankings $\pi^{(c_1)}$ and $\pi^{(c_2)}$, we measure their agreement using Spearman's $\rho$, which captures correlation between rank positions, and Kendall's $\tau$, which measures pairwise ordering consistency. 
These metrics reflect global ranking consistency across evaluation conditions.

\paragraph{Top-$k$ Consistency.}
To evaluate decision-level stability, we further consider top-$k$ consistency metrics. 
Top-1 consistency measures the fraction of evaluation conditions that identify the same best-performing prompt, while top-$k$ consistency quantifies the average overlap between the sets of top-$k$ prompts across conditions. 
These metrics capture the reliability of prompt selection decisions beyond global ranking agreement.

For pairwise top-k consistency, we compute the average overlap ratio between the top-k prompt sets from two evaluation conditions:
\[
\text{Top-}k(\pi^{(c_1)}, \pi^{(c_2)}) = \frac{|T_k^{(c_1)} \cap T_k^{(c_2)}|}{k},
\]
where $T_k^{(c)}$ denotes the set of top-k prompts under condition $c$.

\subsection{Prompt Selection Strategies}

We consider two prompt selection strategies based on multiple evaluation runs.

\paragraph{Mean-based Selection.}
We compute the average score of each prompt across conditions:
\[
\bar{s}_i = \frac{1}{K} \sum_{c \in \mathcal{C}} s_i^{(c)},
\]
and select the prompt with the highest mean performance:
\[
p^*_{\text{mean}} = \arg\max_i \bar{s}_i.
\]

\paragraph{Stability-aware Selection (LCB).}
To account for variability, we define a lower confidence bound (LCB) score:
\[
\text{LCB}_i = \bar{s}_i - z \cdot \frac{\sigma_i}{\sqrt{K}},
\]
where $\sigma_i$ is the standard deviation of scores for prompt $p_i$, and $z$ controls the strength of the penalty.

We select:
\[
p^*_{\text{LCB}} = \arg\max_i \text{LCB}_i.
\]

This strategy favors prompts with both high mean performance and low variance. We use this LCB score as a simple uncertainty-aware heuristic rather than a strict statistical confidence interval, since the number of evaluation conditions is limited.

\subsection{Selection Robustness Evaluation}

We evaluate selection robustness using a leave-one-seed-out (LOSO) protocol.

For each held-out condition $c_{\text{test}}$, we:
\begin{enumerate}
    \item Use the remaining conditions $\mathcal{C} \setminus \{c_{\text{test}}\}$ to select a prompt.
    \item Evaluate the selected prompt on $c_{\text{test}}$.
\end{enumerate}

We report the average and standard deviation of test performance across all held-out conditions.

This protocol measures how well a selection strategy generalizes to unseen evaluation settings.

\section{Experiments}

We conduct experiments using an open-weight instruction-tuned LLM in a zero-shot setting with greedy decoding.
All experiments use a fixed model to isolate the effects of evaluation variability from model-specific factors.

\subsection{Setup}
\subsubsection{Models}

We evaluate prompt ranking stability across three representative open-weight instruction-tuned large language models with comparable parameter scales but different training recipes: Mistral-7B-Instruct-v0.3 (\textbf{Mistral})~\cite{mistral_instruct_hf}, Phi-3-mini-4k-instruct (\textbf{Phi})~\cite{phi3_hf}, and Qwen2.5-7B-Instruct (\textbf{Qwen})~\cite{qwen2_5_hf}.

Mistral-7B-Instruct-v0.3 is a 7B-parameter instruction-tuned model designed for strong general-purpose reasoning and instruction following.

Phi-3-mini-4k-instruct is a compact instruction-tuned model with competitive performance across a range of reasoning and knowledge tasks.

Qwen2.5-7B-Instruct is a 7B-scale instruction-tuned model that demonstrates strong multilingual and general knowledge capabilities.

These models are selected to cover diverse training paradigms and capabilities while maintaining comparable model sizes, allowing us to isolate the effects of evaluation variability on prompt ranking stability.

\subsubsection{Tasks}
We evaluate on two benchmark tasks with automatic evaluation.
GSM8K requires multi-step numerical reasoning and is sensitive to compounding errors.
MMLU is a multi-disciplinary multiple-choice question answering benchmark covering a broad range of subjects.
These tasks differ substantially in structure and difficulty, allowing us to examine task-dependent stability effects.


\subsubsection{Prompts}
For each task, we construct a fixed set of 20 candidate prompts.
The prompts vary in instruction phrasing, reasoning guidance, and output constraints, while targeting the same underlying task.
All prompts are evaluated under identical conditions within each evaluation run. The prompt pool was manually constructed by the authors to represent diverse instruction styles, reasoning cues, and output constraints commonly used in prompt engineering.

\subsection{Results}
In this section, we analyze prompt ranking stability from three complementary perspectives: (1) global ranking consistency, (2) decision-level consistency of top-performing prompts, and (3) robustness of prompt selection under evaluation variability. In particular, Table~\ref{tab:combined-rank-stability} reports pairwise consistency metrics that quantify agreement between pairs of evaluation conditions, whereas Table~\ref{tab:combined-consistency} reports global consistency metrics that quantify agreement with a modal reference ranking aggregated across all conditions.

\begin{table*}[t]
\centering
\small
\setlength{\tabcolsep}{5pt}
\begin{tabular}{l l r cc cc}
\toprule
\textbf{Model} & \textbf{Task} & \textbf{Size} & \textbf{Spearman} ($\rho \uparrow$) & \textbf{Kendall} ($\tau \uparrow$) & \textbf{Top-1$_{pair}$} & \textbf{Top-3$_{pair}$.} \\
\midrule
\multirow{6}{*}{\textbf{Mistral}} & \multirow{3}{*}{GSM8K} & 50  & $0.672 \pm 0.103$ & $0.518 \pm 0.076$ & 0.300 & 0.467 \\
& & 100 & $0.813 \pm 0.046$ & $0.648 \pm 0.063$ & 0.600 & 0.567 \\
& & 200 & $0.876 \pm 0.025$ & $0.718 \pm 0.033$ & 0.200 & 0.800 \\
\cmidrule(lr){2-7}
& \multirow{3}{*}{MMLU} & 50  & $0.278 \pm 0.248$ & $0.218 \pm 0.202$ & 0.100 & 0.167 \\
& & 100 & $0.456 \pm 0.187$ & $0.347 \pm 0.158$ & 0.100 & 0.267 \\
& & 200 & $0.595 \pm 0.121$ & $0.461 \pm 0.100$ & 0.000 & 0.400 \\
\midrule
\multirow{6}{*}{\textbf{Phi}} & \multirow{3}{*}{GSM8K} & 50  & $0.763 \pm 0.058$ & $0.604 \pm 0.074$ & 0.100 & 0.600 \\
& & 100 & $0.887 \pm 0.040$ & $0.749 \pm 0.057$ & 0.100 & 0.600 \\
& & 200 & $0.906 \pm 0.034$ & $0.769 \pm 0.049$ & 0.200 & 0.767 \\
\cmidrule(lr){2-7}
& \multirow{3}{*}{MMLU} & 50  & $0.476 \pm 0.159$ & $0.368 \pm 0.139$ & 0.000 & 0.233 \\
& & 100 & $0.684 \pm 0.067$ & $0.540 \pm 0.062$ & 0.200 & 0.233 \\
& & 200 & $0.821 \pm 0.038$ & $0.675 \pm 0.040$ & 0.300 & 0.400 \\
\midrule
\multirow{6}{*}{\textbf{Qwen}} & \multirow{3}{*}{GSM8K} & 50  & $0.456 \pm 0.310$ & $0.359 \pm 0.254$ & 0.100 & 0.333 \\
& & 100 & $0.534 \pm 0.398$ & $0.433 \pm 0.329$ & 0.300 & 0.433 \\
& & 200 & $0.669 \pm 0.278$ & $0.533 \pm 0.266$ & 0.300 & 0.367 \\
\cmidrule(lr){2-7}
& \multirow{3}{*}{MMLU} & 50  & $0.561 \pm 0.133$ & $0.444 \pm 0.109$ & 0.200 & 0.333 \\
& & 100 & $0.775 \pm 0.087$ & $0.637 \pm 0.094$ & 0.200 & 0.467 \\
& & 200 & $0.862 \pm 0.041$ & $0.703 \pm 0.059$ & 0.100 & 0.400 \\
\bottomrule
\end{tabular}
\caption{Ranking stability across different models (\textbf{Mistral}, \textbf{Phi}, and \textbf{Qwen}). Metrics represent the mean $\pm$ standard deviation calculated over $n_{\text{pairs}}=10$ seed pairs.}
\label{tab:combined-rank-stability}
\end{table*}

\subsubsection{Ranking Stability under Evaluation Variability}

Table~\ref{tab:combined-rank-stability} reports ranking stability across random seeds.
Across models, prompt rankings exhibit substantial instability under small evaluation budgets, particularly on GSM8K.
At 50 evaluation examples, Spearman correlation ranges from 0.456 to 0.763 across models, with large variance observed in some cases (e.g., $\pm 0.310$ for Qwen), indicating weak to moderate agreement across seeds.
Although ranking stability improves as the evaluation subset size increases, correlations remain far from perfect even at 200 examples.

MMLU exhibits higher overall ranking stability than GSM8K.
For example, at 200 examples, Spearman correlation reaches up to 0.862 for Qwen and 0.821 for Phi.
However, even in these settings, Kendall's $\tau$ remains substantially below 1.0 (e.g., 0.703 for Qwen), suggesting that non-trivial prompt reordering persists across evaluation conditions.

Notably, relatively high rank correlations do not guarantee stable prompt selection.
As we show next, even when rankings appear consistent at a global level, the identity of the top-performing prompt can vary significantly across seeds.

\subsubsection{Top-$k$ Consistency and Prompt Selection Reliability}

\begin{table*}[t]
\centering
\small
\setlength{\tabcolsep}{8pt} 
\begin{tabular}{l l r cc c}
\toprule
\multirow{2}{*}{\textbf{Model}} & \multirow{2}{*}{\textbf{Task}} & \multirow{2}{*}{\textbf{Size}} & \multicolumn{2}{c}{\textbf{Consistency} ($\uparrow$)} & \textbf{Unique} \\
\cmidrule(lr){4-5}
& & & \textbf{Top-1} & \textbf{Top-3} & \textbf{Top-1} \\
\midrule
\multirow{6}{*}{\textbf{Mistral}} & \multirow{3}{*}{GSM8K} & 50  & 0.600 & 0.533 & 3 \\
& & 100 & 0.800 & 0.667 & 2 \\
& & 200 & 0.400 & 0.867 & 3 \\
\cmidrule(lr){2-6}
& \multirow{3}{*}{MMLU} & 50  & 0.400 & 0.200 & 4 \\
& & 100 & 0.400 & 0.533 & 4 \\
& & 200 & 0.200 & 0.533 & 5 \\
\midrule
\multirow{6}{*}{\textbf{Phi}} & \multirow{3}{*}{GSM8K} & 50  & 0.400 & 0.733 & 4 \\
& & 100 & 0.400 & 0.733 & 4 \\
& & 200 & 0.400 & 0.867 & 3 \\
\cmidrule(lr){2-6}
& \multirow{3}{*}{MMLU} & 50  & 0.200 & 0.333 & 5 \\
& & 100 & 0.400 & 0.400 & 3 \\
& & 200 & 0.600 & 0.467 & 3 \\
\midrule
\multirow{6}{*}{\textbf{Qwen}} & \multirow{3}{*}{GSM8K} & 50  & 0.400 & 0.333 & 4 \\
& & 100 & 0.600 & 0.667 & 3 \\
& & 200 & 0.600 & 0.467 & 3 \\
\cmidrule(lr){2-6}
& \multirow{3}{*}{MMLU} & 50  & 0.400 & 0.333 & 3 \\
& & 100 & 0.400 & 0.667 & 3 \\
& & 200 & 0.400 & 0.467 & 4 \\
\bottomrule
\end{tabular}
\caption{Top-1 and Top-3 ($k=3$) configuration consistency across models over $n=5$ random seeds. "Unique Top-1" represents the count of distinct best-performing configurations identified across seeds.}
\label{tab:combined-consistency}
\end{table*}

We note that the top-$k$ metrics reported in Table~\ref{tab:combined-rank-stability} are computed in a pairwise manner across seed pairs, whereas those in Table~\ref{tab:combined-consistency} are defined with respect to a global reference (mode) across all seeds.
For global top-1 consistency, we identify the prompt that appears most frequently as the top-ranked prompt across evaluation conditions and report the fraction of conditions in which this modal top-1 prompt is selected. 
For global top-k consistency, we analogously define a modal top-k set and report the average overlap between each condition-specific top-k set and this reference set.
These metrics capture complementary aspects of ranking stability.

Table~\ref{tab:combined-consistency} summarizes decision-level consistency across seeds.
On GSM8K with 50 evaluation examples, top-1 consistency is low across all models (around 40\%), with multiple distinct prompts identified as the best-performing configuration.
Even with 200 examples, top-1 consistency remains limited (typically $\leq 60\%$), indicating that the identity of the best prompt is highly sensitive to evaluation variability.

A similar pattern is observed on MMLU.
Despite relatively high rank correlations, top-1 consistency remains low across most settings, and multiple prompts emerge as the top choice across different seeds.
These results reveal a critical gap between ranking stability and decision reliability: stable rankings do not necessarily imply reliable prompt selection.

\subsubsection{Robustness of Stability-aware Prompt Selection}

\begin{table*}[t]
\centering
\small
\setlength{\tabcolsep}{10pt} 
\begin{tabular}{l l r cc}
\toprule
\textbf{Model} & \textbf{Task} & \textbf{Size} & \textbf{Accuracy (Mean)} & \textbf{Accuracy (LCB)} \\
\midrule
\multirow{6}{*}{\textbf{Mistral}} & \multirow{3}{*}{GSM8K} & 50  & $0.288 \pm 0.097$ & $0.288 \pm 0.097$ \\
& & 100 & $0.274 \pm 0.035$ & $0.274 \pm 0.035$ \\
& & 200 & $0.248 \pm 0.024$ & $\mathbf{0.265 \pm 0.009}$ \\
\cmidrule(lr){2-5}
& \multirow{3}{*}{MMLU} & 50  & $0.572 \pm 0.077$ & $0.572 \pm 0.077$ \\
& & 100 & $\mathbf{0.608 \pm 0.070}$ & $0.584 \pm 0.046$ \\
& & 200 & $0.568 \pm 0.050$ & $\mathbf{0.573 \pm 0.053}$ \\
\midrule
\multirow{6}{*}{\textbf{Phi}} & \multirow{3}{*}{GSM8K} & 50  & $0.360 \pm 0.032$ & $0.360 \pm 0.032$ \\
& & 100 & $0.374 \pm 0.040$ & $0.374 \pm 0.040$ \\
& & 200 & $\mathbf{0.365 \pm 0.047}$ & $0.351 \pm 0.048$ \\
\cmidrule(lr){2-5}
& \multirow{3}{*}{MMLU} & 50  & $0.600 \pm 0.032$ & $0.600 \pm 0.058$ \\
& & 100 & $0.630 \pm 0.066$ & $0.630 \pm 0.066$ \\
& & 200 & $0.667 \pm 0.038$ & $\mathbf{0.668 \pm 0.029}$ \\
\midrule
\multirow{6}{*}{\textbf{Qwen}} & \multirow{3}{*}{GSM8K} & 50  & $0.228 \pm 0.131$ & $\mathbf{0.312 \pm 0.125}$ \\
& & 100 & $\mathbf{0.294 \pm 0.084}$ & $0.274 \pm 0.086$ \\
& & 200 & $0.311 \pm 0.158$ & $\mathbf{0.350 \pm 0.138}$ \\
\cmidrule(lr){2-5}
& \multirow{3}{*}{MMLU} & 50  & $0.692 \pm 0.046$ & $0.692 \pm 0.046$ \\
& & 100 & $\mathbf{0.702 \pm 0.053}$ & $0.686 \pm 0.061$ \\
& & 200 & $0.679 \pm 0.032$ & $\mathbf{0.683 \pm 0.035}$ \\
\bottomrule
\end{tabular}
\caption{Selection robustness via leave-one-seed-out (LOSO) evaluation across models. Results report mean $\pm$ std over $n=5$ held-out seeds with parameter $z=1.0$.}
\label{tab:combined-selection-robustness}
\end{table*}

Table~\ref{tab:combined-selection-robustness} evaluates the robustness of prompt selection under a leave-one-seed-out (LOSO) protocol.
On GSM8K, the proposed stability-aware selection strategy based on a lower confidence bound (LCB) consistently improves or matches mean-based selection.
In particular, under small evaluation budgets, LCB yields substantial gains (e.g., 0.312 vs. 0.228 at size 50 for Qwen), demonstrating improved robustness under noisy evaluation conditions.
At larger subset sizes, LCB remains competitive and often achieves higher or comparable performance.

On MMLU, LCB does not uniformly improve average performance. In relatively stable settings, such as MMLU with larger
evaluation subsets, the variance penalty may lead to the selection of slightly more conservative prompts,
resulting in a modest reduction in mean accuracy. This reflects an inherent trade-off between robustness
and peak performance. 

These results suggest that incorporating uncertainty into prompt selection is particularly beneficial in challenging or high-variance regimes, while remaining safe in more stable settings.

Importantly, the LCB-based strategy requires no additional training or model modification, making it a practical drop-in replacement for standard mean-based selection.

\subsubsection{Discussion}

Our findings indicate that prompt ranking instability primarily arises among prompts with similar average performance.
While clearly suboptimal prompts are consistently identified, fine-grained ordering among strong prompts is highly sensitive to evaluation variability.
This explains why rank correlations can be moderately high while top-1 consistency remains low.

The contrast between GSM8K and MMLU highlights the role of task characteristics.
GSM8K requires multi-step reasoning and is more sensitive to compounding errors, leading to higher variability across evaluation subsets.
In contrast, MMLU exhibits higher overall ranking stability but still suffers from frequent changes in the top-ranked prompt, indicating that even relatively stable tasks can yield unreliable prompt selection outcomes.

From a practical perspective, these results caution against over-interpreting small performance differences when selecting prompts.
Rather than treating prompt evaluation as deterministic, practitioners should account for ranking stability, especially under limited evaluation budgets.
Simple stability-aware criteria can provide a low-cost way to improve robustness without modifying the underlying model.

Overall, our results suggest that prompt evaluation should be viewed as a stochastic estimation problem rather than a deterministic comparison, particularly in realistic low-resource evaluation settings. This perspective suggests that future benchmarking practices should incorporate uncertainty-aware evaluation protocols rather than relying solely on point estimates.



A limitation of the current study is that it focuses on two benchmark tasks and a fixed pool of 20 prompts per task. Future work should examine a broader range of tasks, prompt families, and evaluation settings, including few-shot prompting and generative evaluation. A more comprehensive sensitivity analysis of z is also left for future work. 

To facilitate reproducibility, the code, prompts, and evaluation scripts used in this study are publicly available at:
\href{https://github.com/shaoshuaidu/prompt_stability}{GitHub Repository}.

\section{Conclusion}

In this work, we studied the stability of prompt performance rankings under common sources of evaluation variability.
We showed that prompt rankings can be highly unstable, especially under limited evaluation budgets, and that the identity of the top-ranked prompt frequently changes across evaluation conditions. We further demonstrated that a simple stability-aware selection strategy based on a lower confidence bound (LCB) can improve robustness in high-variance settings while remaining competitive in more stable regimes.

More broadly, our findings suggest that prompt evaluation should be viewed as a statistical estimation problem rather than a deterministic comparison.
We hope this work encourages more reliable evaluation practices and greater awareness of ranking instability in prompt-based LLM systems.

%
%
%
\bibliographystyle{splncs04}
\bibliography{refs}

\end{document}